\pdfoutput=1

\documentclass[11pt]{article}

\usepackage[preprint]{acl}

\usepackage{times}
\usepackage{latexsym}
\usepackage[T1]{fontenc}
\usepackage[utf8]{inputenc}
\usepackage{microtype}
\usepackage{inconsolata}
\usepackage{graphicx}
\usepackage{enumitem}
\usepackage{makecell}
\usepackage{booktabs}
\usepackage{multirow}
\usepackage{subfigure}
\usepackage{framed}
\usepackage{mdframed}

\newtheorem{definition}{Definition}
\usepackage{listings}
\DeclareCaptionStyle{ruled}{labelfont=normalfont,labelsep=colon,strut=off} %
\usepackage{xcolor}
\definecolor{lightgray}{rgb}{0.95, 0.95, 0.95}
\title{AgentMove: A Large Language Model based Agentic Framework for Zero-shot Next Location Prediction}

\author{Jie Feng, Yuwei Du, Jie Zhao, Yong Li  \\
  Department of Electronic Engineering, BNRist, Tsinghua University, Beijing, China, \\
  \{fengjie, liyong07\}@tsinghua.edu.cn
  }
  
\begin{document}
\maketitle

\begin{abstract}
Next location prediction plays a crucial role in various real-world applications. Recently, due to the limitation of existing deep learning methods, attempts have been made to apply large language models (LLMs) to zero-shot next location prediction task. However, they directly generate the final output using LLMs without systematic design, which limits the potential of LLMs to uncover complex mobility patterns and underestimates their extensive reserve of global geospatial knowledge. In this paper, we introduce \textbf{AgentMove}, a systematic agentic prediction framework to achieve generalized next location prediction. In AgentMove, we first decompose the mobility prediction task and design specific modules to complete them, including spatial-temporal memory for individual mobility pattern mining, world knowledge generator for modeling the effects of urban structure and collective knowledge extractor for capturing the shared patterns among population. Finally, we combine the results of three modules and conduct a reasoning step to generate the final predictions. Extensive experiments utilizing mobility data from two distinct sources reveal that AgentMove surpasses the leading baseline by 3.33\% to 8.57\% across 8 out of 12 metrics and it shows robust predictions with various LLMs as base and also less geographical bias across cities. Our codes are available via \url{https://github.com/tsinghua-fib-lab/AgentMove}.
\end{abstract}

\section{Introduction} \label{sec:intro}
Mobility prediction is of great importance in many real-world scenarios, e.g., recommending travel services, pre-activating mobile applications for potential usage, seamless switching of cellular network signals and efficient traffic management. Next location prediction is one of the most important task in human mobility prediction. In recent years, deep learning based models~\cite{liu2016predicting, wu2017modeling, feng2018deepmove, yang2020location, yang2022getnext} have been widely applied and have achieved promising results due to their ability to capture the high-order transition dynamics and mining shared mobility patterns among users. However, existing approaches have several key drawbacks. First, the success of deep learning models rely on the collection of large amounts of private mobility data. Second, the trained model are challenging to apply in zero-shot mobility prediction settings. Finally, the prediction accuracy remains limited due to the constrained sequential modelling capability of smaller deep learning models and a lack of deep understanding of commonsense in human daily life and urban structures.

\begin{figure*}
    \centering
    \includegraphics[width=1\textwidth]{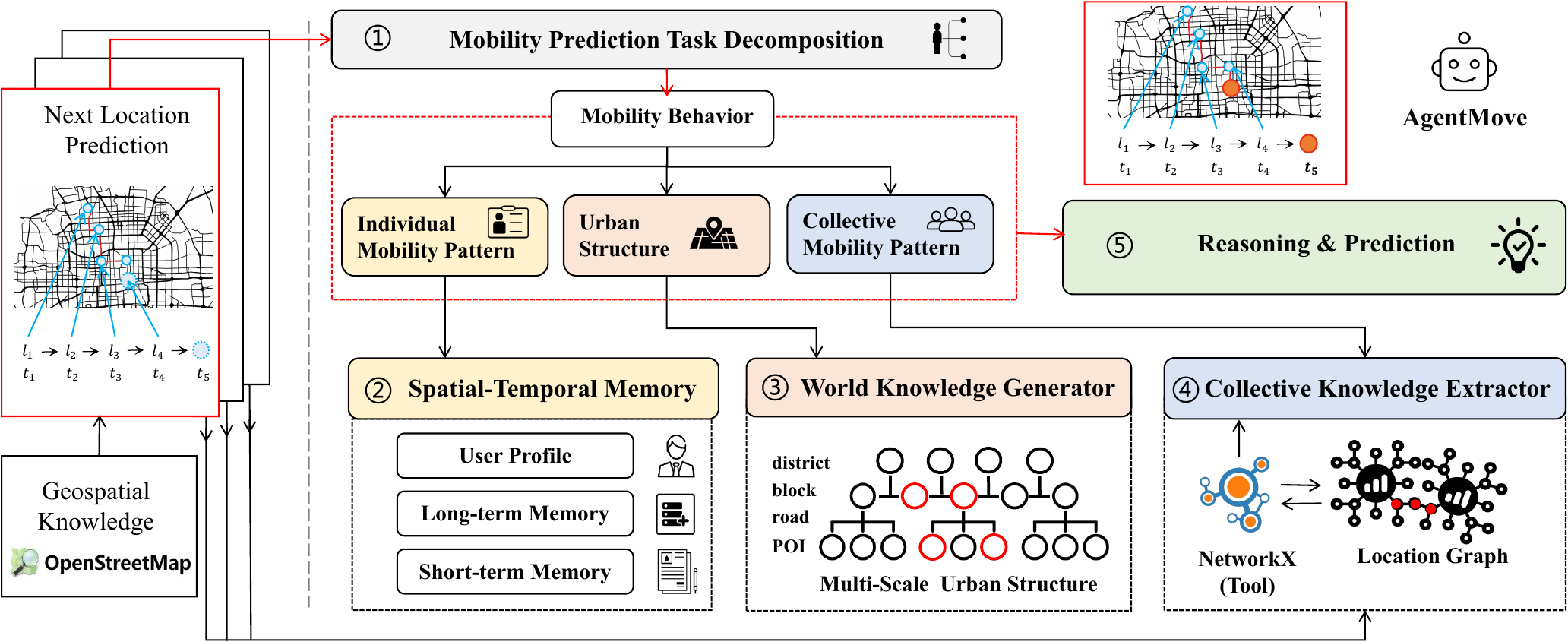}
    \caption{The framework of AgentMove, including three key components: spatial temporal memory unit for capturing individual mobility pattern, world knowledge generator for multi-level urban structure, and collective knowledge extractor for extracting shared mobility patterns among users.}
    \label{figs:framework}
\end{figure*}

Recently, large language models (LLMs) have made significant progress, achieving advanced results that far surpass previous methods in areas such as dialogue-based role-playing, code generation and testing, and mathematics problem solving. In the field of spatial-temporal data mining, researchers are exploring the potential of applying LLMs to various real-world tasks, including time series forecasting~\cite{gruver2024large,li2024urbangpt}, travel planning~\cite{xie2024travelplanner, li2024large}, trajectory analysis~\cite{luo2024deciphering, zhang2023large, du2024trajagent}. Furthermore, several recent works~\cite{wang2023would,beneduce2024large} investigate the feasibility of using LLMs as the base model of mobility prediction, addressing the limitations of deep learning based models and achieving promising results. These works typically convert trajectories to a language based sentence and leverage the powerful sequential modelling capacities of LLMs to directly generate the mobility predictions. However, due to the lack of a systematic design throughout the entire process, they overlook the crucial components of human mobility modeling, resulting in limited performance. In summary, these methods fail to effectively capture the complex individual mobility patterns, neglect to model the effects of urban structure and do not discover the shared mobility patterns among populations.

In this paper, we propose AgentMove, a systematic agentic framework for generalized mobility prediction. By integrating domain knowledge of human mobility, we implement the core components in the general agentic framework~\cite{wang2024survey,xi2023rise}, including the planning module, memory module, world knowledge module, external tool module and reasoning module. For the planning module in AgentMove, we introduce a manually designed mobility prediction task decomposition module that considers the most important factors influencing mobility prediction. This decomposition generates three sub-tasks: individual mobility pattern mining, shared mobility pattern discovery and urban structure modelling. 
First, we implement a spatial-temporal memory module for individual mobility pattern mining. This module contains three submodules--short-term memory, long-term memory and user profiles--to capture the multi-level mobility patterns of individuals. Compared to pure LLM methods, the memory module enables AgentMove to retain past mobility history and efficiently learn from experiences.
Second, we design a world knowledge generator to explicitly extract inherent geospatial knowledge from LLMs, aiding in the modelling the effects of multi-scale urban structures on the human mobility, particularly in relation to the exploration behavior of human mobility.
Third, we equip AgentMove with the capability to discover the shared mobility patterns from various user trajectories through a collective knowledge extractor. This extractor utilizes NetworkX as an external tool to organize trajectories into a global location transition graph and then extract important neighboring locations for prediction. Finally, we combine the results from all the modules and perform a final reasoning step to generate the predictions. In summary, our contributions are as follows,
\begin{itemize}[leftmargin=*]
    \setlength{\itemsep}{0pt}
    \setlength{\parsep}{0pt}
    \setlength{\parskip}{0pt}
    \item To the best of our knowledge, this is the first attempt to apply LLM-based agentic framework to the field of mobility prediction. We build an effective mobility prediction framework by incorporating the crucial characteristics of human mobility into the design of core components.
    \item In AgentMove, we design a spatial-temporal memory module for individual mobility pattern mining, a world knowledge generator for modeling effects of urban structures, and a graph based collective knowledge extractor for discovering the shared mobility patterns among populations. 
    \item Extensive experiments on mobility trajectories from two sources in 12 cities demonstrate the effectiveness of proposed AgentMove, which outperforms the best baseline, achieving performance improvements ranging from 3.33\% to 8.57\% in most cases. Additionally, AgentMove presents superior adaptability to different LLMs, as well as greater stability and reduced bias in prediction results across various cities worldwide.
\end{itemize}

\section{Preliminaries} \label{sec:pre}
We define the mobility prediction task and related concepts for use in the following section.
\begin{figure}
    \centering
    \includegraphics[width=0.45\textwidth]{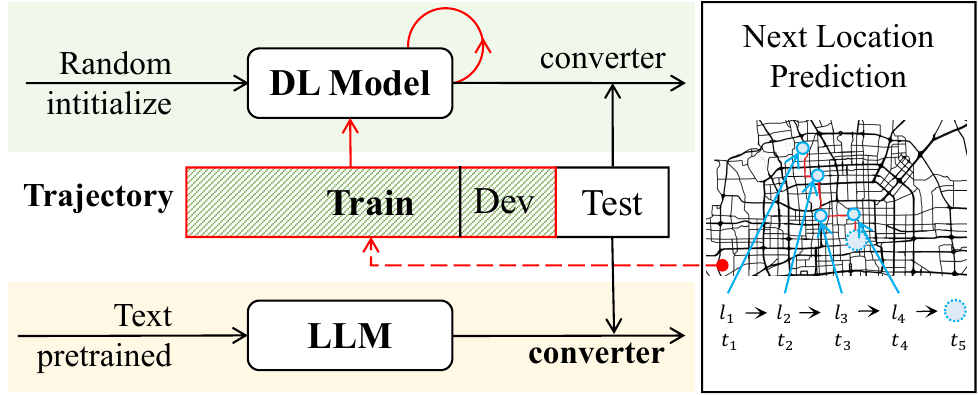}
    \caption{Deep learning and LLM-based mobility predictors work in different ways. Deep learning models need to learn from training data for specific regions, while LLMs predict directly using zero-shot reasoning with its world knowledge.}
    \label{fig:idea}
\end{figure}

\begin{definition}[Location] A location point $p\in P$ is represented as a tuple $\langle id,cate,lon,lat,addr \rangle$, where $id$ is the unique identifier, $cate$ is the category (e.g., restaurant), $lon$ and $lat$ are the coordinates of the location, $addr$ is the text address of location.
\end{definition}

\begin{definition}[User Trajectory] A trajectory of user $u\in U$ is represented as $T_{u}=\lbrace (p_1, t_1), (p_2, t_2), \dots, (p_n, t_n)\rbrace$, where $p_i \in P$ is the $i$-th location visited by the user and $t_i$ is the timestamp of the visit.
\end{definition}

\begin{definition}[Contextual Stays] Contextual stays of user $u$ is defined as the most recent sub-sequence in trajectory: $\mathcal{C}_u = \lbrace (p_{n-k}, t_{n-k}), \dots, (p_{n-1}, t_{n-1}), (p_n, t_n)\rbrace$, which captures the user's short-term mobility patterns. $k$ is the window size of contextual stays.
\end{definition}

\begin{definition}[Historical Stays]
Historical stays of user $u$ is defined as the sub-sequence before contextual stays: $\mathcal{H}_u=\lbrace (p_{1}, t_{1}),(p_{2}, t_{2}), \dots, (p_{n-k-1}, t_{n-k-1})\rbrace $, which captures the user's long-term mobility patterns.
\end{definition}

Given the historical movement data $\mathcal{C}_u$, $\mathcal{H}_u$ as well as available external knowledge $\mathcal{K}$ (e.g., worldwide geospatial information), the objective is to predict the next location $p_{n+1}$ that user $u$ will visit. Formally, this paper aims to learn a mapping function $f$:
\begin{equation}
    f: (\mathcal{C}_u, \mathcal{H}_u, \mathcal{K}) \rightarrow p_{n+1}.
\end{equation}
Figure~\ref{fig:idea} illustrates the differences between the deep learning based paradigm and the LLM based paradigm in the mobility prediction task. The deep learning model needs collecting training data before conduct the prediction task, which means it cannot directly used in the zero-shot scenario. LLM based method can directly applied into any scenario after carefully `format converter' (known as prompt engineering). While LLM based methods can be adapted easily to new scenarios, their effectiveness may not improve as the scenarios accumulates more data. In this way, the deep learning models with more data can achieve better performance when LLM based methods fail to improving. In this paper, we propose the LLM based agent solution AgentMove for mobility prediction task which enables the continue learning and improving of LLM based mobility predictor.  

\section{Methods} \label{sec:method}

\subsection{Overview}
As shown in Figure~\ref{figs:framework}, AgentMove comprises five core components: task decomposition module, spatial-temporal memory module, world knowledge generator, collective knowledge extractor and the final reasoning module. Serving as the high-level planning module, the task decomposition module is designed to break down the overall mobility prediction task into subtasks---personalized mobility pattern mining, collective mobility pattern discovery and modelling the effects of urban structures---by considering the crucial factors influencing mobility. The detailed design of the other components is introduced as follows.

\subsection{Spatial-temporal Memory}
\begin{figure}
    \centering
    \includegraphics[width=1\columnwidth]{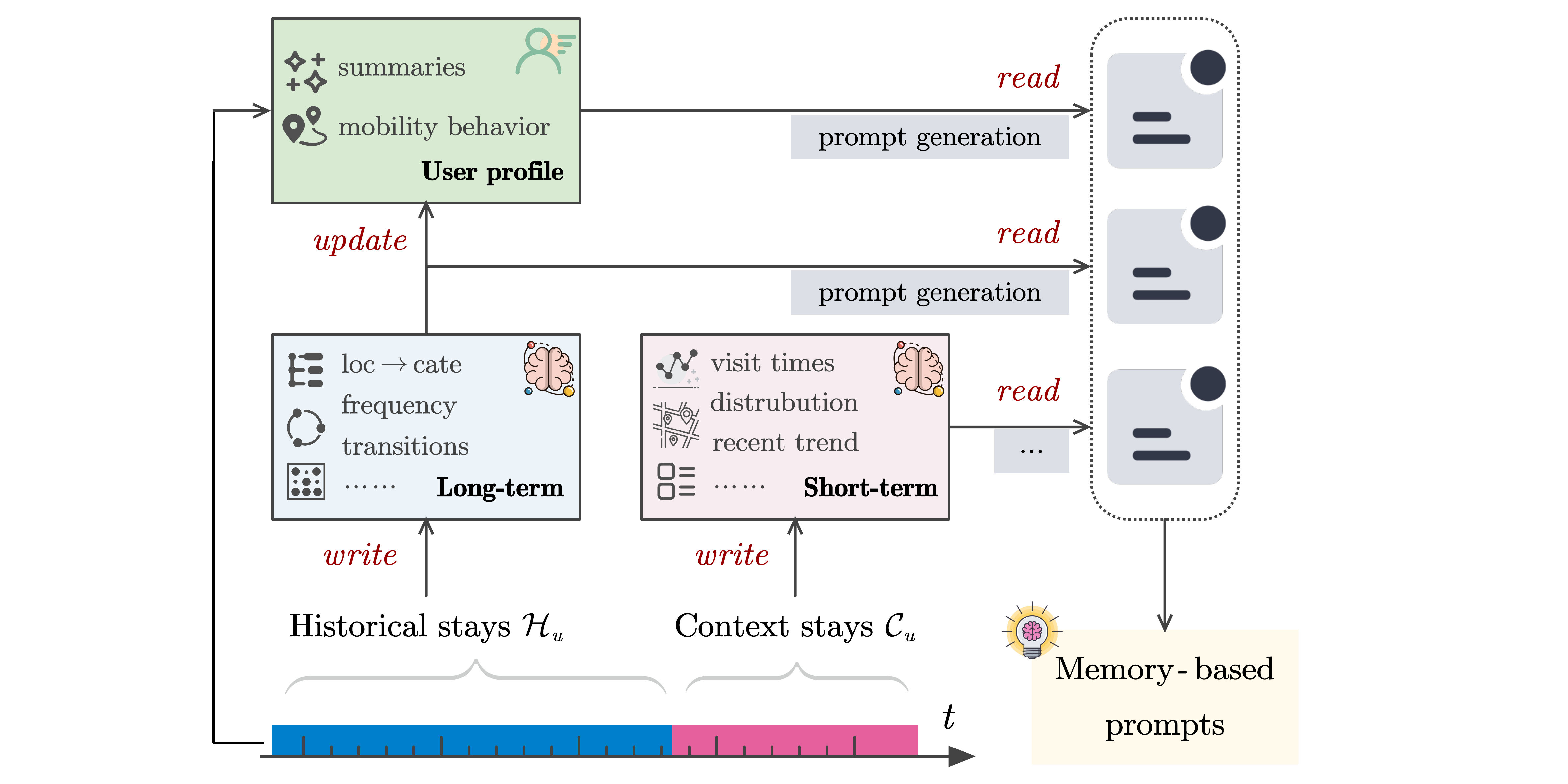}
    \caption{Illustration of spatial-temporal memory.}
    \label{figs:st_mem}
\end{figure}

The spatial-temporal memory module is designed to effectively capture, store and leverage mobility patterns, providing crucial insights for the personalized and multi-scale periodicity behavior modelling in mobility prediction. Inspired by the memory design principles in general LLM-based agents~\cite{zhang2024survey}, our spatial-temporal memory functions through three essential processes: memory organization, memory writing, and memory reading. 
The whole framework of spatial--temporal memory module is presented in Figure~\ref{figs:st_mem}.

\subsubsection{Memory Organization}
The spatial-temporal memory is structured into three components to capture multifaceted nature of user mobility patterns: \textit{User Profile Unit}. This unit provides a summary description of the user's mobility behavior as the user mobile profile, which offers deeper insights into when and why the user visits certain locations. The user profile is dynamically generated based on the current long-term memory introduced in the following part, allowing AgentMove to adapt to the evolving user preferences; \textit{Long-term Memory Unit}. This unit retains users' long-term mobility patterns, capturing overarching trends and recurring sequences in their movement history. It functions similarly to how LLMs store long-term dependencies in textual data; \textit{Short-term Memory Unit}. This unit focuses on users' recent mobility patterns, providing dynamic updates that reflect the latest movements and short-term variations.

All users' memories are stored in a central memory pool, organized as key-value pairs. Each key corresponds to a unique user identifier, and the value consists of the long-term memory, short-term memory, and user profile info. This organization ensures a comprehensive extraction and storage of mobility data, enabling efficient retrieval and utilization for mobility prediction.

\subsubsection{Memory Writing}
Writing to the memory involves the extraction and structured storage of spatial-temporal patterns hidden in user's trajectories. This process consists of two main steps:

\textbf{Long-term Memory Writing}. Given the historical stays $\mathcal{H}_u$, this module extracts long-term spatial-temporal information of user $u\in U$, including:
1) \textit{location to category mapping}. Associating visited locations with their respective categories.
2) \textit{top-$k$ active times and locations}. Identifying the most active time periods and the most frequently visited locations.
3) \textit{location visit frequency}. Recording how often various locations are visited.
4) \textit{transition matrix}. A matrix that represents the transition probabilities between locations.

\textbf{Short-term Memory Writing}. Given the contextual stay $\mathcal{C}_u$, this module extracts fine-grained short-term spatial-temporal information of user $u\in U$, including:
1) \textit{time sequence of recent visits}. Documenting the sequence of recent visit times.
2) \textit{visit frequency of different locations}. Tracking how frequently different locations are visited in the short term.
3) \textit{details of the last visit}. Recording specific details about the latest location visit.

By systematically organizing and storing this information by processing the trajectories,  AgentMove can easily access to both long-term and short-term mobility patterns. This structured approach is crucial for enhancing the accuracy of next location predictions.

\subsubsection{Memory Reading}
The memory reading process involves generating spatial-temporal context relevant prompts from the structured memory to enhance AgentMove's predictive capabilities. This process consists of three key steps:

\textbf{User Profile Prompt Generation}. Utilizing the long-term memory, AgentMove constructs user profile prompts that encapsulate the intrinsic movement patterns and habitual behaviors of the user. These prompts include summaries of peak activity times, preferred locations, and temporal-spatial associations, providing a comprehensive mobility profile of user.

\textbf{Long-term Memory Prompt Generation}. Also based on the long-term memory, AgentMove generates prompts by summarizing the user's general mobility trends from the long-term view. These prompts include details on the most active times, frequently visited locations, and the relationships between these factors. This helps the LLM understand the user's regular movement patterns.

\textbf{Short-term Memory Prompt Generation}. AgentMove creates prompts from the short-term memory to reflect recent mobility patterns and contextual information of user. These prompts cover recent visit sequences, current visit frequencies, and specifics of the latest visits, which ensure LLMs efficiently adapt to recent changes in user's behavior.

Finally, these memory-based prompts are consolidated into a cohesive spatial-temporal summary of the original trajectory, which is then integrated as the first part of AgentMove's prompts. This spatial-temporal summary enhances the LLM’s ability to engage in more logical and efficient reasoning, leading to more precise mobility predictions.

\subsection{World Knowledge Generator}
Numerous studies~\cite{jiang2016timegeo} indicate that individual movement typically encompasses two types of behaviors: returning and exploring. As introduced before, the return behavior has been well-captured by the spatial-temporal memory module. In this section, we introduce the world knowledge generator, which extracts geospatial knowledge from LLMs and constructs a multi-scale urban structure to enable the modelling of explore behavior in mobility. To extract geospatial knowledge effectively, we propose aligning the knowledge of LLMs and the urban space of trajectory via text addresses. Once the spaces are aligned, we explicitly prompt the LLMs to generate potential candidate places for exploration from the perspective of the multi-scale urban structure. 

\subsubsection{Alignment via Address}
Many existing works~\cite{feng2018deepmove,luo2021stan,lin2021pre,cui2021st,qin2022next, hong2023context} on mobility prediction usually represent the locations directly using latitude and longitude coordinates or discrete spatial area IDs. While this approach facilitates the easy construction of deep learning-based spatial encoding, it is not suitable for LLMs. Since LLMs are trained on large scale human-generated text, they, like human, are not inherently adept at understanding the precise coordinates~\cite{gurnee2023language} or discrete area IDs. Thus, we propose to utilize the text address which human is familiar to describe the coarse location of trajectory. While text address is not precise as the coordinates, it is more natural and easy to be aligned with the existing spatial knowledge in the LLMs. 

Thus, we adapt the address searching service~\footnote{https://nominatim.org/} from Open Street Map to build address for each point in the trajectory. Due to cultural and institutional differences, address information formats vary greatly across different countries. To address this, we leverage the common-sense knowledge of LLMs to extract unified structured address information from the original address information. LLMs can easily pinpoint a user’s current and past locations, laying a solid foundation for subsequent modeling.
\subsubsection{Multi-scale Urban Structure}
Based on the real structured address information, we design prompts to motivate LLMs to generate multi-scale potential places which may be explored by user in the future. We introduce multi-scale generation mechanism to help LLMs reduce hallucination and improve the accuracy and usability of generate places. The multi-scale location information covers four level: district, block, street and POI name. We first ask LLMs to generate the potential districts in the future. Then, based on these districts and the past blocks in the trajectory to generate the potential blocks in the future and so on. Finally, we can generate potential location information from different levels as the potential exploration candidate for the user. 

\subsection{Collective Knowledge Extractor}
In the previous two sections, we introduce the spatial-temporal memory module and world knowledge generator for the individual-level mobility modelling. Here, we introduce the collective knowledge extractor, which captures shared mobility patterns among users to further enhance the mobility predictions. First, we construct a global location transition graph using NetworkX~\footnote{https://networkx.org/documentation/stable/index.html} by aggregating the location transitions from various users. We then employ a LLM to perform simple reasoning on the graph, utilizing functions in NetworkX as tool to generate potential locations visited by other users with similar mobility patterns. 

\subsubsection{Building Location Transition Graph}
In the location graph, the node is location ID with various attributes, e.g., address information, function of location. The edge between nodes is constructed by considering the 1-hop transition between nearby trajectory points in each trajectory. The edge is weighted without direction. Based on the definition of graph, we use NetWorkX to build the graph from scratch and update it when infer trajectories for various users. If any history trajectory data, e.g. training data used by the deep learning based models, are available, the location graph can be initialized by them. 

\subsubsection{Reasoning on Graph}
After obtaining the location graph, we can utilize LLM to perform reasoning on the whole graph via the function of NetworkX as tool. The most naive strategy is to query the k-hop neighbors of the current location. When the number of the neighbors is too much, LLMs need to filter the most promising ones from them by considering the attributes of each node. Furthermore, we can extend the query nodes into the last n locations and generate the most promising ones from all the neighbors of them. In this way, we obtain the most relevant locations that has been visited by the users with similar mobility patterns. 

\subsection{Summarization and Prediction}
Finally, we design prompts to employ LLM to analyze and summarize the information from different views and perform a final reasoning step to generate the prediction with reasons. The prompts for output format requirements are also be placed here to ensure that the output format meets the requirements as much as possible. Detailed prompts can refer to the appendix.

\section{Evaluation} \label{sec:eval}

\begin{table*}
\centering
\setlength{\tabcolsep}{1mm}
\resizebox{1\textwidth}{!}{
\begin{tabular}{lcccccccccccc}
\toprule
 \textbf{Model} & \multicolumn{3}{c}{\textbf{FSQ@Tokyo}}& \multicolumn{3}{c}{\textbf{FSQ@SaoPaulo}}& \multicolumn{3}{c}{\textbf{FSQ@Moscow}}&  \multicolumn{3}{c}{\textbf{ISP@Shanghai}}\\
  & \textbf{Acc@1} & \textbf{Acc@5}  & \textbf{NDCG@5} & \textbf{Acc@1} & \textbf{Acc@5}  & \textbf{NDCG@5} & \textbf{Acc@1} & \textbf{Acc@5}  & \textbf{NDCG@5} &  \textbf{Acc@1}& \textbf{Acc@5}& \textbf{NDCG@5}\\ 
\midrule
  \textbf{FPMC} & 0.060& 0.165& 0.121& 0.045& 0.085& 0.066& 0.020& 0.065& 0.043 & 0.13& 0.355&0.249\\
  \textbf{RNN} & 0.105& 0.240& 0.176& 0.095& 0.230& 0.169& 0.090& 0.185& 0.140 & 0.065& 0.175&0.123\\
  \textbf{DeepMove} & 0.175& 0.320& 0.251& 0.150& 0.310& 0.236& \underline{0.165}& 0.335& 0.258& \underline{0.175}& 0.320&0.251\\ 
  \textbf{LSTPM} & 0.145& 0.280& 0.218& 0.190& 0.365& 0.281& 0.140& 0.255& 0.196 & 0.095& 0.17&0.135\\
 \textbf{GETNext}& \textbf{0.205}& \underline{0.450}& \underline{0.317}& 0.165& 0.375& 0.258& 0.175& \textbf{0.380}& \textbf{0.269}& 0.115& 0.260&0.178\\
 \textbf{STHGCN}& \underline{0.198}& 0.430& 0.300& 0.175& \underline{0.398}& 0.299& \textbf{0.180}& \underline{0.372}& 0.265& 0.125& 0.277&0.195\\ 
\midrule
 \textbf{LLM-Mob} & 0.175& 0.370& 0.277& 0.140& 0.275& 0.210& 0.080& 0.175& 0.129 & 0.100& 0.345&0.221\\
  \textbf{LLM-ZS} & 0.175& 0.410& 0.299& 0.165& 0.385& 0.277& 0.120& 0.340& 0.233 & 0.170& \underline{0.425}&0.298\\
 \textbf{LLM-Move}& 0.145& 0.285& 0.243& \underline{0.220}& 0.355& \underline{0.325}& 0.155& 0.270& 0.226& 0.140& 0.410&\underline{0.308}\\ 
\midrule
 \textbf{AgentMove} & 0.185& \textbf{0.465}& \textbf{0.331}& \textbf{0.230}& \textbf{0.415}& \textbf{0.326}& 0.160& 0.365& \underline{0.265}& \textbf{0.190}& \textbf{0.450}&\textbf{0.329}\\
 \textbf{vs Deep Learning}& -9.76\%& 3.33\%& 4.42\%& 25.71\%& 4.27\%& 9.03\%& -11.11\%& -3.95\%& -1.49\%& 8.57\%& 40.63\%&31.08\%\\
 \textbf{vs Best Baselines}& -9.76\%& 3.33\%& 4.42\%& 4.55\%& 4.27\%& 0.31\%& -11.11\%& -3.95\%& -1.49\%& 8.57\%& 5.88\%&6.82\%\\
\bottomrule
\end{tabular}
}
\caption{The main results of baselines and AgentMove. GPT4omini is used as the base LLM for all the LLM-based methods in the table. Deep learning methods are first trained on the training set of each city and LLM-based models are directly evaluated on the test set with the zero-shot prediction settings.}
\label{table:main}
\end{table*}

\subsection{Settings}
\subsubsection{Datasets} \label{sec:data}
We use the global Foursquare \textbf{checkin data}~\cite{yang2016participatory}  and recent public released ISP \textbf{GPS trajectory data}~\cite{feng2019dplink} to conduct the experiments. The Foursquare data contains checkins from 415 cities which covers about 18 months from April 2012 to September 2013. The ISP GPS trajectory data is from the mobile network logs in Shanghai with 325215 records, covering April 19 to April 26 in 2016. Compared with Foursquare data, ISP data is much denser and was open-sourced in June 2024~\footnote{https://github.com/vonfeng/DPLink/tree/master/data}, which is beyond the training period of all the LLMs used in the experiment. This ensures that the evaluation results are not affected by potential data leakage.

To evaluate the general mobility prediction ability of AgentMove, we select 12 cities from the Foursquare dataset and the entire ISP trajectory data to conduct the experiments. We follow the preprocessing procedure~\cite{hong2023context, feng2019dplink}  to process the trajectories data. For Foursquare checkin data, we divide each trajectory dataset into training, validation, and test sets in a ratio of 7:1:2. While the ISP data lasts only 7 days, we split the whole data into training set, validation set and testing data in a ratio of 4:1:5 for preserving enough testing data. Detailed description about preprocessing can refer to the appendix. We follow the data license in the original paper and use these trajectory data only for academic purpose.

We select Tokyo, Moscow and SaoPaulo with the largest amount of Foursquare check-in data and the ISP data from Shanghai to conduct the main analysis in the experiments and results of 12 cities are discussed in the final section of experiment. 
We divide each trajectory dataset into training, validation, and test sets. The training and validation sets are only used to train the deep learning model, and the resulting models are compared with the LLM-based methods on the test set. 
Due to the cost of the various API calling, e.g.,  Llama3.1-405B,  we randomly sample 200 instances from the testing set for each city to calculate the performance in the experiments.

\subsubsection{Baselines}
We compare proposed models with following baselines: \textbf{FPMC}~\cite{rendle2010factorizing}, five deep learning models (\textbf{RNN}~\cite{feng2018deepmove}, \textbf{DeepMove}~\cite{feng2018deepmove}, \textbf{LSTPM}~\cite{sun2020go}, \textbf{GETNext}~\cite{yang2022getnext}, \textbf{STHGCN}~\cite{yan2023spatio}) and three LLM-based methods(\textbf{LLM-Mob}~\cite{wang2023would}, \textbf{LLM-ZS}~\cite{beneduce2024large}, \textbf{LLM-Move}~\cite{feng2024move}). We use widely used Accuracy@1, Accuracy@5, and NDCG@5 as the main evaluation metrics~\cite{sun2020go, luca2021survey} in the experiments.

\subsubsection{Implementation}
We use LibCity~\cite{jiang2023libcity} to implement the FPMC, RNN, DeepMove and LSTPM. We use the official codes from author to implement GETNext~\footnote{https://github.com/songyangme/GETNext} and STHGCN\footnote{https://github.com/ant-research/Spatio-Temporal-Hypergraph-Model}. We follow the default parameter settings of these models in the library and official codes for training and inference. For LLMs, we use OpenAI API~\footnote{https://platform.openai.com/} for accessing GPT4omini, DeepInfra~\footnote{https://deepinfra.com/} and SiliconFlow~\footnote{https://siliconflow.cn/models} for accessing other open source LLMs. Detailed parameter settings for those baselines can be found in the appendix.

\subsection{Main Results}

In this section, we compare AgentMove with 9 baselines in 4 cities at Table~\ref{table:main}. In the experiments, we use GPT4omini as the default base LLM for all LLM-based methods.

As the representative deep learning models, GETNEext and STHGCN achieve best or second-best results in 4 out of 12 metrics. Compared with the deep learning baselines, the best LLM-based baseline LLM-Move can achieve better results than GETNext and STHGCN in 3 out of 12 metrics, which present the powerful sequential pattern discovery and reasoning ability of LLM in modeling mobility. It is noted that the results of LLM-based methods are zero-shot prediction while the deep learning based methods rely on sufficient training with enough mobility data. Compared with these baselines, our proposed method AgentMove is the best method and achieves the best results in 8 out of 12 metrics in 4 datasets. Although AgentMove falls slightly behind the best baseline, GETNext, in three metrics, two of them are very close. 
These results in Table~\ref{table:main} demonstrate the effectiveness of proposed framework in stimulating the comprehensive ability of LLM-based agentic framework for mobility prediction.

\subsection{Ablation Study on Model Designs}
In this section, we provide a more detailed analysis of the proposed method under varying model designs to further demonstrate its effectiveness.

We first conduct ablation study to demonstrate the contribution of each component in AgentMove for its excellent performance, which are presented in Table~\ref{table:ablation}. We first discuss the impact of three core components individually, as detailed in the top four lines of Table~\ref{table:ablation}. Overall, all components contribute to performance improvement in most cases. However, the performance gains vary across different metrics. For example, while memory design leads to better performance in the Acc@1 in SaoPaulo, the performance in other three metrics are dropped. The effects of the combination of the core components in the last three lines in Table~\ref{table:ablation}. In summary, compared with the base prompt design, the combination of proposed designs introduce 7\%-45\% performance gain in all the datasets.

Besides, to demonstrate the effects of the World Knowledge Generator (WKG) in exploring new locations, we analyze whether our model explores more potential locations that are not present in the user's recent contextual stays after incorporating the WKG module. The results are presented in the Table~\ref{table:wkg}. A higher percentage indicates that the model tends to revisit locations from the recent contextual stays, while a lower percentage suggests that the model explores more new locations. The results demonstrate that the WKG successfully encourages the model to explore new locations, which is particularly beneficial for improving performance.

\begin{table}
\centering
\setlength{\tabcolsep}{0.6mm}
\resizebox{0.48\textwidth}{!}{
\begin{tabular}{lcccccc} 
\toprule
\textbf{Models}& \multicolumn{3}{c}{\textbf{FSQ@SaoPaulo }}&  \multicolumn{3}{c}{\textbf{ISP@Shanghai}}\\
 & \textbf{Acc@1} & \textbf{Acc@5} & \textbf{NDCG@5} & \textbf{Acc@1} & \textbf{Acc@5} &\textbf{NDCG@5}  \\ 
\hline
\textbf{base} & 0.165~ & 0.385~ & 0.277~ & 0.170& 0.425&0.298\\
\textbf{+STM} & 0.190~ & 0.315~ & 0.255~ & 0.170& 0.445&0.312\\
\textbf{+WKG} & 0.175~ & 0.365~ & 0.269~ & 0.155& 0.390&0.276\\
\textbf{+CKE} & 0.175~ & 0.380~ & 0.275~ & 0.175& \textbf{0.465}&0.317\\
\textbf{+STM+WKG} & \textbf{0.240~} & \underline{0.390~} & \underline{0.317}~ & \textbf{0.215}& \underline{0.455}&\textbf{0.342}\\
\textbf{AgentMove} & \underline{0.230~} & \textbf{0.415~} & \textbf{0.326~} & \underline{0.190}& 0.450&\underline{0.329}\\
\textbf{vs base} & 45.45\% & 7.79\% & 17.99\% & 11.76\%& 5.88\%&10.30\%\\
\bottomrule
\end{tabular}}
\caption{Ablation studies of AgentMove. `base' denotes the basic prompts which is similar to the baseline LLM-ZS, `+STM' denotes adding spatial-temporal memory, `+WKG' denotes adding world knowledge generator, `+CKE' denotes adding collective knowledge extractor.}
\label{table:ablation}
\end{table}

\begin{table}
\centering
\setlength{\tabcolsep}{0.6mm}
\resizebox{0.45\textwidth}{!}{
    \begin{tabular}{lccc} 
    \toprule
    \multicolumn{2}{l}{Location return rate} & FSQ@SaoPaulo$\downarrow$ & ISP@Shanghai$\downarrow$ \\ 
    \midrule
    \multirow{2}{*}{LLama3-8b} & w/ WKG & 94\%& \textbf{75.6\%}\\
     & w/o WKG & \textbf{93\%}& 87.8\%\\ 
    \midrule
    \multirow{2}{*}{LLama3-70b} & w/ WKG & \textbf{87.5\%}& \textbf{73.2\%}\\
     & w/o WKG & 90\%& 85.4\%\\
    \bottomrule
    \end{tabular}}
    \caption{Effectiveness of word knowledge generator (WKG) for encouraging mobility exploration, which is measured by the location return rates. The location return rate measures the tendency to revisit previously visited locations based on recent contextual stays.}
    \label{table:wkg}
\end{table}

\subsection{Geographical Bias and LLM Effects} \label{sec:bias}
\begin{figure}
    \centering
    \subfigure[Acc@1 of three LLM-based methods on 12 cities.]{
        \includegraphics[width=0.22\textwidth]{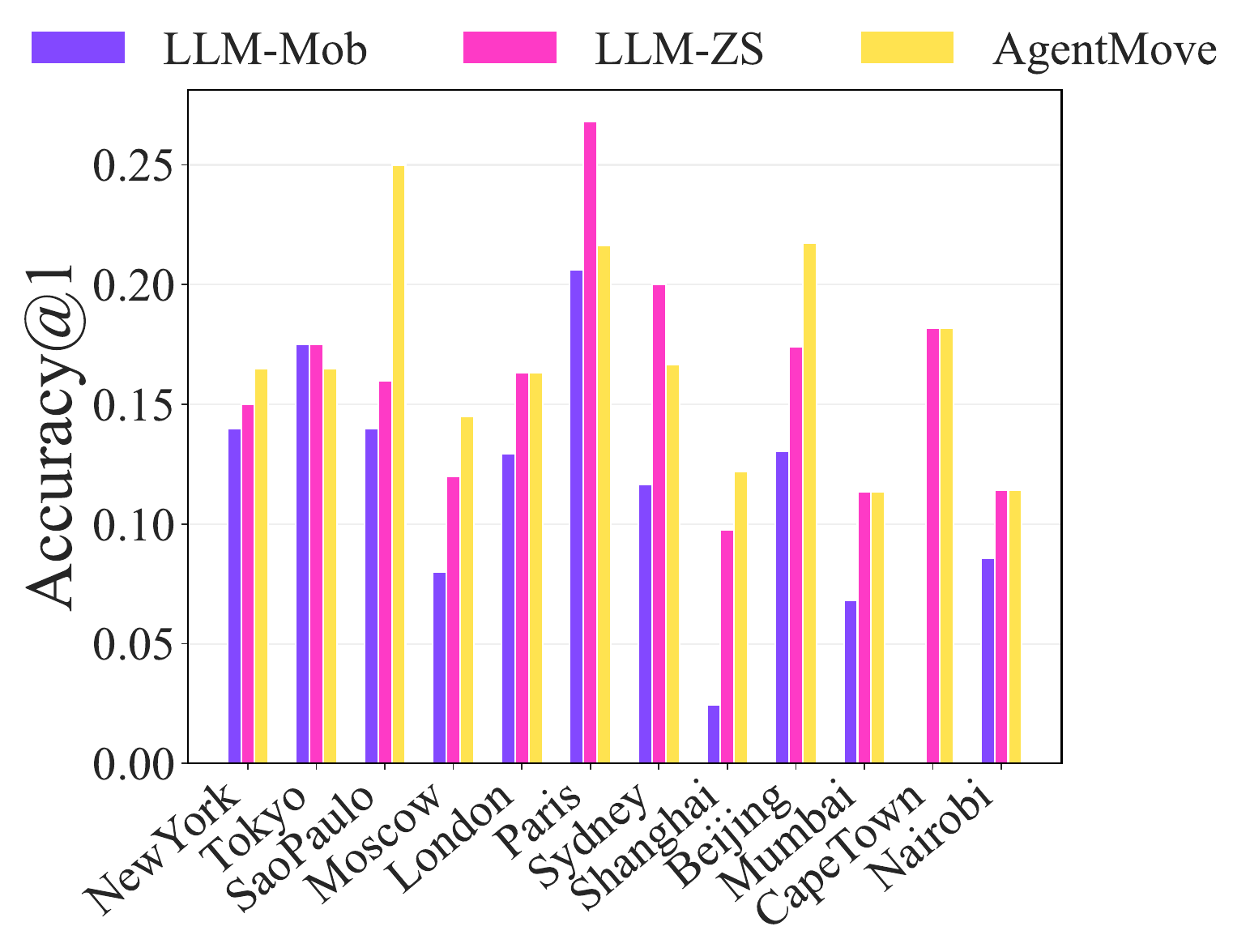}
        \label{figs:geo:methods}
    }
    \subfigure[Distribution of Acc@5 across 3 methods on 12 cities.]{
        \includegraphics[width=0.21\textwidth]{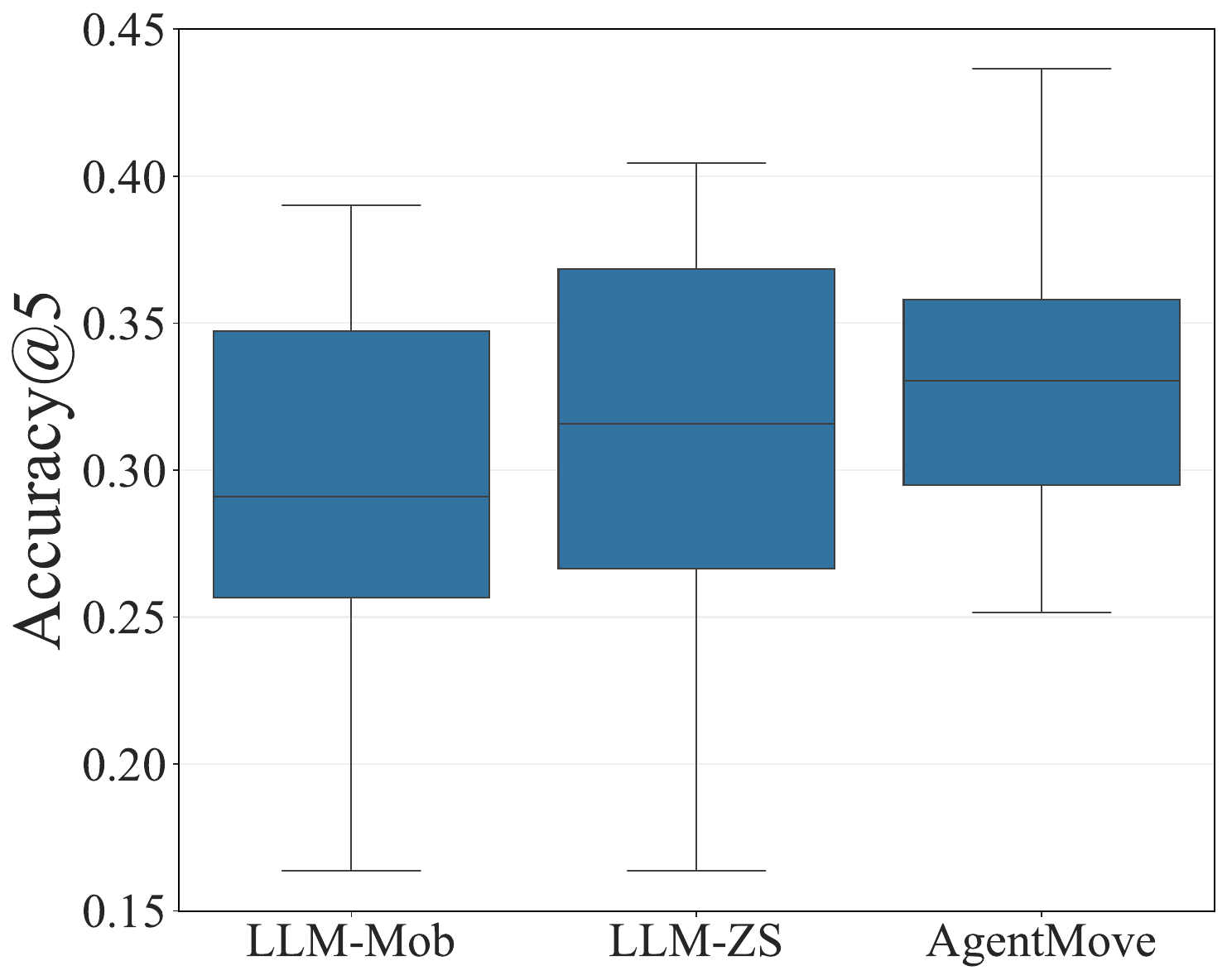}
        \label{figs:geo:acc5}
    }
    
    \caption{Geospatial bias analysis of various methods in mobility prediction across 12 cities, where AgentMove outperforms most methods and exhibits lower geospatial bias.}
    \label{fig:bias}
\end{figure}

While LLMs are trained with the online web text which can be geographically bias~\cite{manvi2024large} around the world. We investigate the potential geographical bias in LLM based mobility prediction methods and attempt to answer whether AgentMove can alleviate the geographical bias inherent in LLMs to some extent. Experiment results conducted on 12 cities are presented in Figure~\ref{fig:bias}. 

In Figure~\ref{figs:geo:methods}, we can find significant differences in the accuracy of three LMM-based methods across cities. For instance, cities like Tokyo, Paris, and Sydney generally achieve higher accuracy, while cities like Cape Town and Nairobi see notably lower performance. This suggests the presence of geographical bias in trained LLMs. We also find that proposed AgentMove performs best in most of the cities.
Figure~\ref{figs:geo:acc5} provides a box-plot test by comparing the Acc@5 of the three LLM-based methods in 12 cities. Results demonstrate that AgentMove not only outperforms the other methods in terms of overall accuracy but also exhibits a smaller range of error. The performance of AgentMove is more consistent across different cities, suggesting a reduced impact of geographical bias with carefully designs in it.

\begin{figure}
    \centering
    \subfigure[Effects of LLM types on three LLM-based methods]{
        \includegraphics[width=0.22\textwidth]{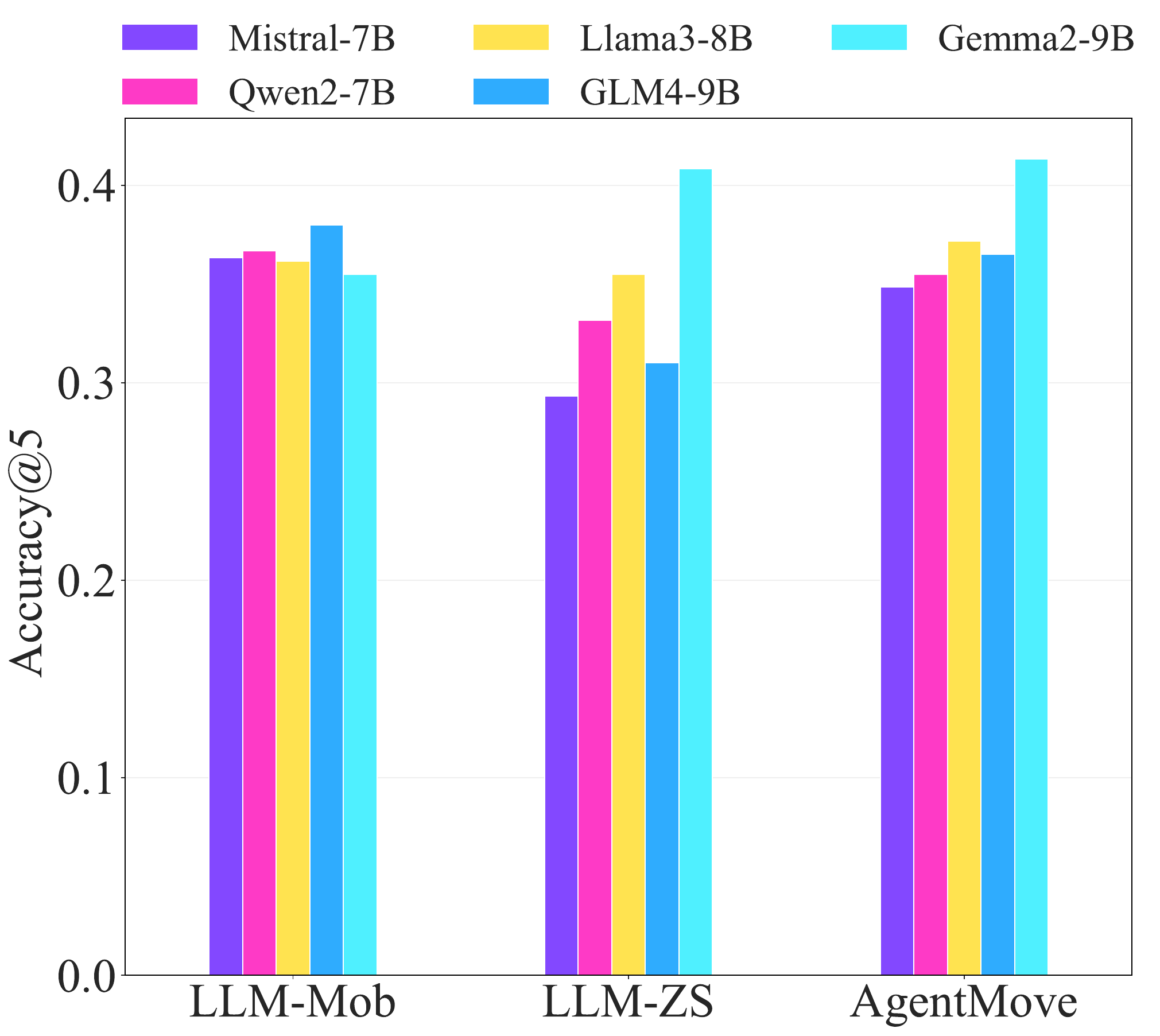}
        \label{figs:llms:source}
    }
    \subfigure[Effects of LLM size on three cities.]{
        \includegraphics[width=0.22\textwidth]{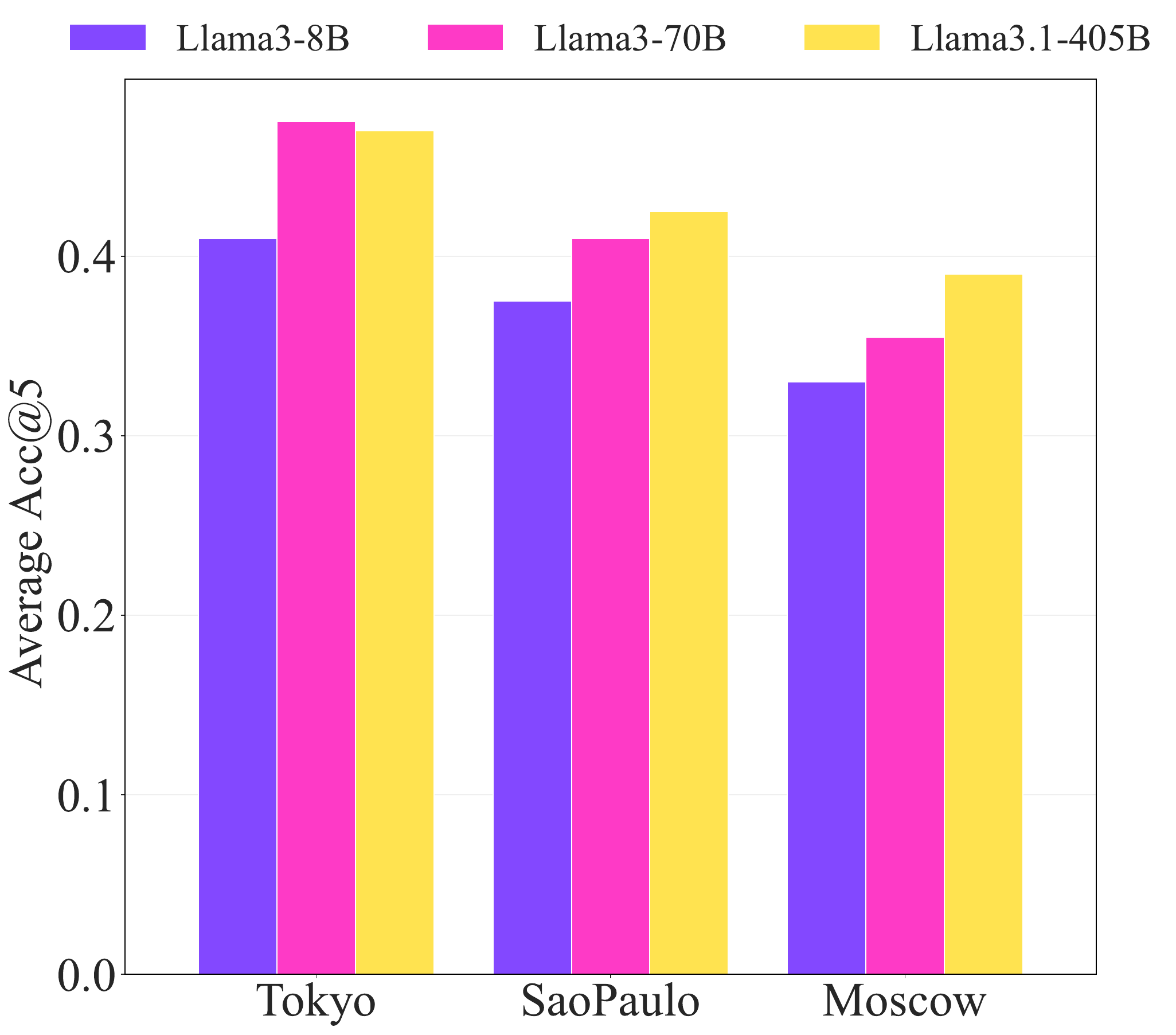}
        \label{figs:llms:cities}
    }
    
    \caption{The effects of LLM with varying sizes and sources on the prediction performance of three LLM based methods.}
    \label{figs:llms}
\end{figure}

As the core foundation of AgentMove, the capabilities of base LLM play a critical role in its performance. Thus, we evaluate the impact of different LLMs with varying sizes and structures in Figure~\ref{figs:llms} by using FSQ@Tokyo. Figure~\ref{figs:llms:source} presents the impact of various 7B LLM with different training data and model structures. The results show that proposed AgentMove performs best adaptability among different LLMs. While LLM-Mob performs stable in all the 7B-LLMs, its performance on Gemma2-9B is far worse than other two methods. 
We then discuss the detailed impacts of LLM size on AgentMove's performance across different data. Figure~\ref{figs:llms:cities} reveals that that larger models, particularly Llama3.1-405B, generally deliver significant performance gains for AgentMove compared to smaller models like Llama3-8B across different cities. It is also observed that in Tokyo, Llama3-1-405B performs slightly weaker compared to Llama3-70B. This suggests that while larger models often excel, their effectiveness may vary depending on the unique mobility patterns and characteristics of each city.

\section{Related Work} \label{sec:related}
\subsection{Mobility Prediction with Deep Learning}
Significant efforts have been made in mobility prediction using deep learning models, encompassing research from both \textit{sequential-based methods} and \textit{graph-based methods}. Traditional approaches typically employ Markov models~\cite{rendle2010factorizing, cheng2013you} to predict the next visit by learning the transition probabilities between consecutive POIs. In contrast, sequential-based deep learning methods have been proposed to model the high-order movement patterns in trajectory data. These methods can be categorized into two types: recurrent neural networks(RNNs)~\cite{kong2018hst, huang2019attention, yang2020location, zhao2020discovering, feng2020pmf}, and attention mechanisms\cite{feng2018deepmove,luo2021stan,lin2021pre,cui2021st,qin2022next, hong2023context} based works. 

Despite their success, these methods primary focus on extracting mobility patterns from an individual perspective, while overlooking the collaborative information available from other users' trajectories. To address this limitation, recent works~\cite{rao2022graph,yang2022getnext} have explored the use of graph neural network(GNNs) for their ability to model complex relationships. 
However, all these methods rely on collecting large volumes of private trajectory data. In contrast, our AgentMove leverages the world knowledge and sequential modeling abilities of LLMs to enable the generalized mobility prediction with zero-shot prediction ability.

\subsection{Large Language Models and Agents}
Due to the powerful language-based generalization and reasoning capabilities~\cite{wei2022emergent}, large language models~\cite{ChatGPT,touvron2023llama} have developed rapidly and have been widely applied in various tasks, including programming~\cite{qian2024scaling} and mathematics~\cite{wei2022emergent}. Recent studies~\cite{gurnee2023language,manvi2023geollm} have found that LLMs possess a significant amount of geographical knowledge about the world. Additionally, researchers also explore the potential of applying LLMs in spatial-temporal data modelling by directly converting domain-specific tasks into a language-based format, such as time series forecasting~\cite{gruver2024large}, traffic prediction~\cite{li2024urbangpt}, trajectory mining~\cite{wang2023would, beneduce2024large}, trip recommendation~\cite{xie2024travelplanner, li2024large}, traffic signal control~\cite{lai2023large, feng2024citybench}, comprehensive urban tasks~\cite{feng2024citygpt, feng2024citybench}. 

These early works highlight the potential of LLMs in spatial-temporal modelling. To effectively utilize the vast knowledge embedded in LLMs and stimulate their reasoning and planning abilities, various prompt techniques~\cite{wei2022chain, kojima2022large, wang2022self, yao2024tree} have been proposed for solving naive text games and mathematical problems. However, when for more complex real-life and domain-specific tasks, these simple prompt techniques alone are insufficient. Recently, LLM based agents~\cite{wang2024survey,xi2023rise,du2024trajagent} are been proposed to address this limitation by equipping LLMs with explicit memory, structured workflows and external tools. In this work, we are the first to design LLM based agent specifically for the mobility prediction task. By incorporating explicit spatial-temporal memory and a workflow for geospatial and social structure mining, we successfully leverage the world knowledge of LLMs and their structured reasoning capabilities for mobility trajectory modelling.

\section{Conclusion}
In this paper, we propose AgentMove, a systematic agentic framework for generalized human mobility prediction applicable worldwide. We design a spatial-temporal memory module and a collective knowledge extractor to learn both individual mobility patterns and shared mobility pattern among users. Furthermore, we develop a world knowledge generator that utilizes text-based address to understand urban structures in a manner similar to humans. Extensive experiments on trajectories from 12 cities demonstrate the superiority and robustness of AgentMove for mobility prediction.

In the future, we plan to explore more effective ways to extract and leverage the vast world knowledge and common sense of LLMs for mobility modeling and trajectory mining. Additionally, we aim to extend the framework to other trajectory data mining tasks, such as trajectory classification and generation. We believe that LLM-based agents, like AgentMove, hold great potential and adaptability, paving the way for a new paradigm in spatial-temporal modeling alongside deep learning.

\section{Limitations}
Here, we discuss the potential limitations of the current work and outline directions for future exploration.

\textbf{Robustness and Hallucination} Based on an LLM, the output of AgentMove is not fully controllable. In this work, we define a simple output parser to extract the expected context from the LLM output, but it may occasionally fail. Due to the potential for hallucination in LLMs, the output of AgentMove may include false addresses that do not exist in the real world. While we can define a clear list of valid locations during experiments to verify this, doing so in real-world applications presents significant challenges.

\textbf{High Cost} The high cost of invoking the LLM API limited our experiments to 12 cities with a small test set. This cost will also pose a challenge for large-scale deployment in real-world scenarios. The reliance on LLMs does pose a significant limitation in terms of scalability of our method. With ongoing advancements~\cite{liu2024mobilellm,qu2025mobile} in the development of more efficient and scalable LLM alternatives—such as smaller LLMs, model pruning, and knowledge distillation—we are optimistic about the potential for rapidly decreasing computational costs while improving scalability. 

\textbf{Geospatial Bias} Geographical bias has long been a challenging issue in LLMs~\cite{manvi2024large}. While our proposed AgentMove incorporates specific design elements to mitigate some of these biases, it cannot completely eliminate them due to inherent limitations in LLMs. However, we believe that our current work represents a significant step forward in addressing geospatial bias in mobility prediction challenges. One promising direction for further reducing geographical bias could be the integration of more external knowledge during inference, and we are actively exploring this avenue in our future work.

\section{Ethics Statement}
All trajectory data used in the experiments come from publicly available open-source datasets~\cite{yang2016participatory, feng2019dplink}. We do not attempt to extract any personal information from these datasets.

\bibliography{bibliography}

\section{Appendix}
\subsection{Baselines}
\begin{itemize}[leftmargin=1.5em,itemsep=0pt,parsep=0.2em,topsep=0.0em,partopsep=0.0em]
    \item \textbf{FPMC}~\cite{rendle2010factorizing} It combines the matrix factorization and Markov chains methods together for sequential modeling. 
    \item \textbf{RNN}~\cite{feng2018deepmove} It is a simple RNN based mobility prediction model as regarding the mobility sequence as general sequence. 
    \item \textbf{DeepMove}~\cite{feng2018deepmove} It contains a LSTM for capturing the short-term sequential transition and an attention unit for extracting long-term periodical patterns.
    \item \textbf{LSTPM}~\cite{sun2020go} It consists of a non-local network for long-term modeling and a geo-dilated RNN for short-term learning.
    \item \textbf{GETNext}~\cite{yang2022getnext} It use transition flow map to assistant a transformer based model to predict next location with cold start settings. 
    \item \textbf{STHGCN}~\cite{yan2023spatio} It designs a novel hypergraph transformer to capture higher-order relations between trajectories for prediction.
    \item \textbf{LLM-Mob}~\cite{wang2023would} It is the first work to apply LLM (GPT-3.5) to predict the next location. 
    \item \textbf{LLM-ZS}~\cite{beneduce2024large} It defines simple prompts and testifies more LLMs in zero-shot mobility prediction task.
    \item \textbf{LLM-Move}~\cite{feng2024move} It uses RAG to provide nearby POIs for LLM to predict next location more precisely. 
\end{itemize}

\subsection{Discussion about the usage of text-based locations}
In most studies on mobility prediction, numerical representations, such as coordinates, are widely used. In this work, we incorporate text-based location information as the main part of the input. While geographic coordinates can precisely describe location information, they lack the semantic context necessary to activate the geospatial knowledge embedded in LLMs. As demonstrated in GeoLLM~\cite{manvi2023geollm}, querying LLMs with raw coordinates alone is often ineffective for tasks like predicting population density. In contrast, text-based representations align naturally with LLMs’ strengths in understanding and reasoning over natural language, allowing them to better leverage their pre-trained spatial knowledge. By converting coordinates into structured text addresses using Open Street Map and LLMs, our approach enriches trajectory points with meaningful geospatial context, such as landmarks and cultural relevance, which raw coordinates cannot provide. This approach strikes a practical balance between precision and contextual richness, optimizing LLMs for human mobility modeling. However, in the future, integrating precise numerical representation with text-based representation presents a promising research direction.

\subsection{Details of Data}
Detailed information of processed trajectory data from 12 cities is presented in Table~\ref{table:data}.

\begin{table}
\centering
\setlength{\tabcolsep}{0.2mm}
\resizebox{0.48\textwidth}{!}{
    \begin{tabular}{cccccc} 
        \toprule
        \textbf{City} & \textbf{Users} & \textbf{Traj.} & \textbf{Loc.} & \textbf{Avg. Traj. }& \textbf{Records} \\ 
        \hline
        \textbf{Tokyo} & 12464 & 112942 & 83190 & 9.06& 1030105 \\ 
        \textbf{SaoPaulo} & 11856 & 77120 & 78904 & 6.50& 809198 \\ 
        \textbf{Moscow} & 10501 & 100854 & 93599 & 9.60& 950898 \\ 
        \textbf{NewYork} & 15785 & 28502 & 41386 & 1.81& 380247 \\ 
        \textbf{Sydney} & 1720 & 4557 & 10523 & 2.65& 54250 \\ 
        \textbf{Paris} & 6903 & 7559 & 19837 & 1.09& 111325 \\ 
        \textbf{London} & 9724 & 14596 & 28687 & 1.50& 188530 \\ 
        \textbf{Beijing} & 1076 & 1847 & 5753 & 1.72& 21030 \\ 
        \textbf{Shanghai-FSQ} & 1272 & 3238 & 8014 & 2.54& 33129 \\
        \textbf{Shanghai-ISP} & 1762 & 2844 & 12576 & 1.61& 325215 \\
        \textbf{Capetown} & 403 & 1234 & 2988 & 3.06& 13303 \\ 
        \textbf{Mumbai} & 1070 & 3070 & 7942 & 2.87& 40592 \\ 
        \textbf{Nairobi} & 356 & 2690 & 5807 & 7.55& 28453 \\
        \bottomrule
    \end{tabular}
}
    \caption{Trajectory statistics of 12 cities around the world.}
    \label{table:data}
\end{table}

\subsection{Examples of Extracted Mobility Behaviors by AgentMove}
\begin{lstlisting}
###### 1. mobility behaviors from spatial-temporal memory#######

## The personal profile and long memory:
<historical_info>: 
- place id to name mapping: '''{'4f58467xxx': 'Middle Eastern Restaurant', '4b058793fxxx': 'Miscellaneous Shop', '4ebaaccfxxx': 'Residential Building (Apartment / Condo)', ....}.'''
- In historical stays, The user frequently engages in activities at 7 AM (2 times), 12 PM (2 times), 4 PM (2 times), ..... 
- The most frequently visited venues are Home (private) (2 times), Middle Eastern Restaurant (1 times), Miscellaneous Shop (1 times), ..... 
- Hourly venue activities include 12 PM: Indian Restaurant (1 times), 2 PM: Home (private) (1 times), 3 PM: Thai Restaurant (1 times), ....
- The user activity transitions often include sequences: '''Middle Eastern Restaurant -> Miscellaneous Shop (1 times), Miscellaneous Shop -> Residential Building (Apartment / Condo) (1 times), ....'''

<user_profile>: 
The user is most active at 7 AM with 2 visits. They frequently visit Home (private) with 2 visits.Based on the data, the user enjoys trying different types of food and drinks.
##############end##################


###### 2. mobility behaviors from world knowledge generator#######
## The potential places from the global spatial view:
### Names of subdistricts that are relatively likely to be visited:
1. Taiyanggong
2. Sanlitun
3. Jiaodaokou Subdistrict
4. Xiaoguan
5. Qianmen
### Names of POIs that are relatively likely to be visited:
1. Yuan Yang Future Plaza Shopping Mall
2. Peking Hostel
3. University of International Business and Economics
4. Beijing Public Library
5. Peking University
##############end##################


########### 3. mobility behaviors from collective knowledge extractor
## The nearby places visited by other users with similar mobility pattern:
1-hop neighbor places in the social world: Xibahu Road, Mars Garden
##############end##################
\end{lstlisting}

\subsection{Prompt Examples}
Here, we present the detailed prompts for each LLM based methods.

\textbf{Prompt of AgentMove} \label{sec:prompt-example}
\begin{lstlisting}
## Task
Your task is to predict <next_place_id> in <target_stay>, a location with an unknown ID, while temporal data is available.

## Predict <next_place_id> by considering:
1. The user's activity trends gleaned from <historical_stays> and the current activities from  <context_stays>.
2. Temporal details (start_time and day_of_week) of the target stay, crucial for understanding activity variations.
3. The potential places that users may visit based on an overall analysis of multi-level urban spaces.
4. The personal profile and memory info extracted from the long trajectory history of each user.

## The potential places from the global spatial view:
{spatial_world_info}

## The nearby places visited by other users with similar mobility pattern:
{social_world_info}

## The personal profile and long memory:
{spatial_temporal_memory_info}

## The history data:
<historical_stays>: {historical_stays}
<context_stays>: {context_stays}
<target_stay>: {target_time, <next_place_id>}

## Output 
Present your answer in a JSON object with:
"prediction" (list of IDs of the five most probable places, ranked by probability) and "reason" (a concise justification for your prediction).
\end{lstlisting}

Prompt for spatial-temporal memory unit.
\begin{lstlisting}
### long term memory info
Place id to name mapping: {venue_id_to_name}.
In historical stays, The user frequently engages in activities at {frequent_hours}.
The most frequently visited venues are {frequent_venues}.
Hourly venue activities include {hourly_activity_desc}.
The user's activity transitions often include sequences such as {transitions}.

### short term memory info
In recent context stays, user's last visit was on {}
Frequently visited locations include: {}
Visit times: {}

### user profile
The user is most active at {most_frequent_hour} with {most_frequent_count} visits.
They frequently visit {most_frequent_venue_category} with {most_frequent_venue_count} visits
Based on the data, the user {', '.join(insights)}.
\end{lstlisting}

Prompts for world knowledge generator.
\begin{lstlisting}
# Prompts for world knowledge generator

## prompts for extracting structured address info
{original address info from https://nominatim.org/ by querying via reverse API}
Please get the administrative area name, subdistrict name/neighbourhood name, access road or feeder road name, building name/POI name.
Present your answer in a JSON object with:'administrative' (the administrative area name) ,'subdistrict' (subdistrict name/neighbourhood name),'poi'(building name/POI name),'street'(access road or feeder road name which POI/building is on).
Do not include the key if information is not given.Do not output other content.

### block info
This trajectory moves within following administrative areas:
{administrative_area}
This trajectory sequentially visited following subdistricts, with the last subdistrict being the most recently visited:{}
Consider about following two aspects:
1.The frequency each subdistrict is visited.
2.Transition probability between two administrative areas.
Please predict the next subdistrict in the trajectory. Give {explore_num} subdistricts that are relatively likely to be visited. Do not output other content.

### poi and street info
This trajectory sequentially visited following POIs(Each POI is represented by 'POI name, the feeder road or access road it is on'), with the last POI being the most recently visited:{pois})
Consider about following two aspects:
1.The frequency each subdistrict is visited.
2.The frequency each poi is visited.
3.Transition probability between two subdistricts.
4.Transition probability between two pois.
Please predict the next poi in the trajectory.Give {explore_num} POIs that are relatively likely to be visited. Do not output other content.


# spatial world model info used in AgentMove
### Names of subdistricts that are relatively likely to be visited:
{block_info}
### Names of POIs that are relatively likely to be visited:
{poi_info}

\end{lstlisting}

Prompt for collective knowledge extractor.
\begin{lstlisting}
## Finding neighbors
neighbors = list(graph.neighbors(venue_id))
sorted_neighbors_freq = [(n, 1) for n in neighbors if n not in context_trajs]

## Prompts in final reasoning step
1-hop neighbor places in the social world: {neighbors}
......
\end{lstlisting}

\textbf{Prompt of LLM-Mob}
\begin{lstlisting}
Your task is to predict a user's next location based on his/her activity pattern.
You will be provided with <history> which is a list containing this user's historical stays, then <context> which provide contextual information 
about where and when this user has been to recently. Stays in both <history> and <context> are in chronological order.
Each stay takes on such form as (start_time, day_of_week, duration, place_id). The detailed explanation of each element is as follows:
start_time: the start time of the stay in 12h clock format.
day_of_week: indicating the day of the week.
duration: an integer indicating the duration (in minute) of each stay. Note that this will be None in the <target_stay> introduced later.
place_id: an integer representing the unique place ID, which indicates where the stay is.

Then you need to do next location prediction on <target_stay> which is the prediction target with unknown place ID denoted as <next_place_id> and 
unknown duration denoted as None, while temporal information is provided.      

Please infer what the <next_place_id> might be (please output the 10 most likely places which are ranked in descending order in terms of probability), considering the following aspects:
1. the activity pattern of this user that you learned from <history>, e.g., repeated visits to certain places during certain times;
2. the context stays in <context>, which provide more recent activities of this user; 
3. the temporal information (i.e., start_time and day_of_week) of target stay, which is important because people's activity varies during different time (e.g., nighttime versus daytime)
and on different days (e.g., weekday versus weekend).

Please organize your answer in a JSON object containing following keys:
"prediction" (the ID of the five most probable places in descending order of probability) and "reason" (a concise explanation that supports your prediction). Do not include line breaks in your output.

The data are as follows:
<historical>: {historical_stays}
<context>: {context_stays}
<target_stay>: {target_time, <next_place_id>}
\end{lstlisting}

\textbf{Prompt of LLM-ZS}
\begin{lstlisting}
Your task is to predict <next_place_id> in <target_stay>, a location with an unknown ID, while temporal data is available.

Predict <next_place_id> by considering:
1. The user's activity trends gleaned from <historical_stays> and the current activities from  <context_stays>.
2. Temporal details (start_time and day_of_week) of the target stay, crucial for understanding activity variations.

Present your answer in a JSON object with:
"prediction" (IDs of the five most probable places, ranked by probability) and "reason" (a concise justification for your prediction).
    
The data:
<historical_stays>: {historical_stays}
<context_stays>: {context_stays}
<target_stay>: {target_time, <next_place_id>}
\end{lstlisting}

\subsection{Parameter settings}
Detailed parameter settings for each Markov and deep learning based baselines are presented in Table~\ref{table:para}. For each baseline, we adapt the early stopping methods by considering the accuracy of validation set and learning rate schedule threshold. All the experiments of deep learning baselines are running on a machine with 64 cores, 512GB of memory, and 2 NVIDIA RTX 4090 GPU, which is installed with Ubuntu 22.04.3 LTS.
\begin{table}
\caption{Detailed parameter settings for Markov and deep learning based baselines.}
\label{table:para}
\centering
\setlength{\tabcolsep}{0.3mm}
\resizebox{0.48\textwidth}{!}{
\begin{tabular}{lcccc} 
\toprule
\textbf{Parameters} & \textbf{FPMC} & \textbf{RNN} & \textbf{DeepMove} & \textbf{LSTPM} \\ 
\hline
\textbf{batch size} & 1024 & 1024 & 128 & 128 \\
\textbf{learning rate (lr)} & - & 1e-3 & 1e-3 & 1e-3 \\
\textbf{lr schedule step} & - & 2 & 3 & 2 \\
\textbf{lr schedule decay} & - & 0.1 & 0.1 & 0.1 \\
\textbf{schedule threshold} & - & 1e-3 & 1e-3 & 1e-3 \\
\textbf{early stop lr} & - & 9e-6 & 9e-6 & 9e-6 \\
\textbf{L2~} & - & 1e-5 & 1e-5 & 1e-6 \\
\textbf{max epoch} & 100 & 30 & 30 & 30 \\
\textbf{loc embed size} & 64 & 500 & 500 & 500 \\
\textbf{hidden embed size} & - & 500 & 500 & 500 \\
\textbf{dropout} & - & 0.3 & 0.5 & 0.8 \\
\bottomrule
\end{tabular}
}
\end{table}

All the generation parameter settings for LLM based methods are the same. The temperature is set as 0 for deterministic results, the maximum output token is 1000, the maximum input token is 2000, other parameters are not set and follow the default settings from API provider.

\subsection{Preprocessing for Foursquare Data} \label{sec:app:data}
As introduced in section~\ref{sec:data}, we select 12 cities around the world to evaluate the performance of proposed framework. We match each trajectories with the target cities by calculating the minimum distance to the city center. For the ordered trajectories in each city, we use 72 hours as the time window to split the trajectory into sessions. We filter the users with less than 5 sessions and filter sessions with less than 4 stays. Then, we divide each trajectory dataset into training, validation, and test sets in a ratio of 7:1:2. During the testing, we filter the users with less than 3 sessions or more than 50 sessions which is designed to ensure the quality of testing users and also balance the effects from different users. Different from the previous works, we do not specifically filter locations. All the users and trajectories of them are sorted by the id. We select one session of each user and aggregate the first $n$ sessions from all the users to calculate the average accuracy. Here, $n$ is utilized to control the cost of evaluation for LLMs and keep fixed in the experiment, which is set as 200 in most of the experiments. It is noted that only the volume of testing set is controlled for cost, the entire training set is provided to the deep learning based methods for training. 
\subsection{Preprocessing for ISP Data}
Following the preprocessing in the original paper~\cite{feng2019dplink}, we split the data into different sessions by merging trajectory points in the same day. Due to the regularity of human, there are too much repeated trajectory points in the original sessions. To make the prediction challenging, we compress the trajectory sessions by merging the same locations within a time window (2 hours) and ignoring the visiting occurred during the night (from 8 p.m. to 8 a.m.). While the ISP data lasts only 7 days, we split the whole data into training set, validation set and testing data in a ratio of 4:1:5 for preserving enough testing data. The minimum session filter parameter is changed from 3 to 1.

\end{document}